\documentclass{article}

\PassOptionsToPackage{numbers,compress}{natbib}

\usepackage[preprint]{latentomni}

\usepackage[utf8]{inputenc} 
\usepackage[T1]{fontenc}    
\usepackage{hyperref}       
\usepackage{url}            
\usepackage{booktabs}       
\usepackage{amsfonts}       
\usepackage{nicefrac}       
\usepackage{microtype}      
\usepackage[table]{xcolor} 
\usepackage{mathtools}
\usepackage{tcolorbox}
\usepackage{tabularx}
\tcbuselibrary{skins, breakable}
\definecolor{myPurple}{RGB}{113, 0, 113}

\newtcolorbox{promptbox}[2][]{
  enhanced,
  breakable,
  colback=white,
  colframe=myPurple,
  coltitle=white,
  colbacktitle=myPurple,
  fonttitle=\bfseries,
  title={#2},
  boxed title style={
    colback=myPurple,
    colframe=myPurple,
    rounded corners
  },
  attach boxed title to top left={
    xshift=0.6cm,
    yshift=-2mm
  },
  top=5mm,
  left=4mm,
  right=4mm,
  bottom=3mm,
  arc=2mm,
  boxrule=1.2pt,
  #1
}
\DeclareRobustCommand{\titleicon}{%
  \raisebox{-0.12\height}{\includegraphics[height=1.5em]{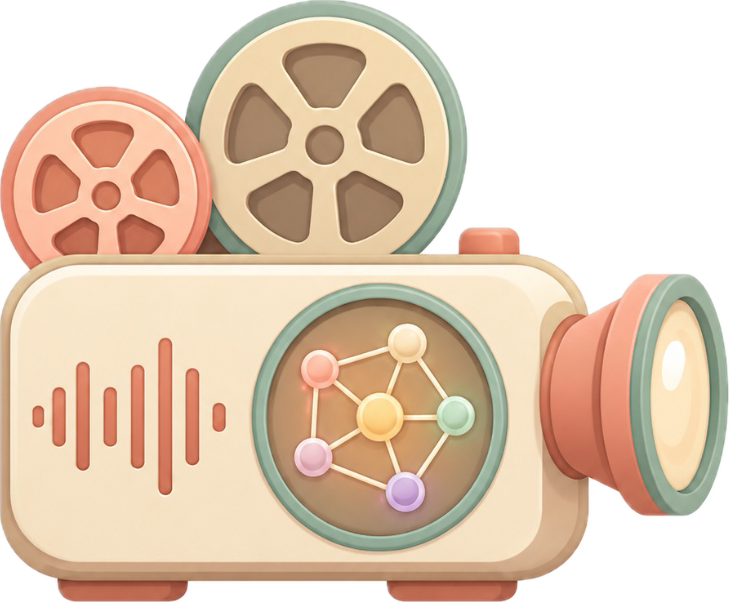}}%
}
\title{
\texorpdfstring{%
{\Large
\titleicon\hspace{0.15em}
LatentOmni: Rethinking Omni-Modal Understanding\\[-0.2em]
via Unified Audio-Visual Latent Reasoning
}%
}{%
LatentOmni: Rethinking Omni-Modal Understanding via Unified Audio-Visual Latent Reasoning
}%
}

\author{
Yifan Dai$^{1,2}$,
Zhenhua Wu$^2$,
Bohan Zeng$^{3,2}$,
Daili Hua$^3$, 
Jialing Liu$^7$,
Bozhou Li$^{3,2}$, \\
\textbf{Yuran Wang}$^{3,2}$,
\textbf{Chengzhuo Tong}$^{3,2}$, 
\textbf{Hao Liang}$^3$,
\textbf{Xiaochen Ma}$^4$, 
\textbf{Junbo Niu}$^3$, \\
\textbf{Tianyu Guo}$^3$,
\textbf{Yang Shi}$^{3,2}$,
\textbf{Yue Ding}$^{5,2}$,
\textbf{Yiyan Ji}$^{6,2}$,
\textbf{Bingyin Mei}$^8$, \\
\textbf{Yushuo Guan}$^2$,
\textbf{Yuanxing Zhang}$^2$,
\textbf{Pengfei Wan}$^2$,
\textbf{Fangcheng Fu}$^1$, 
\textbf{Wentao Zhang}$^3$\\
\small
$^1$School of AI, Shanghai Jiao Tong University,
$^2$Kling Team, Kuaishou Technology,
$^3$Peking University, \\
\small
$^4$HKUST, 
$^5$CASIA,
$^6$Nanjing University,
$^7$Renmin University of China,
$^8$Tsinghua University \\
}

\begin{document}

\maketitle

\begin{abstract}

Joint audio-visual reasoning is essential for omnimodal understanding, yet current multimodal large language models (MLLMs) still struggle when reasoning requires fine-grained evidence from both modalities. A central limitation is that explicit text-based chain-of-thought (CoT) compresses continuous audio-visual signals into discrete tokens, weakening temporal grounding and shifting intermediate reasoning toward language priors. We argue that a unified latent space is a better medium for such reasoning because it preserves dense sensory information while remaining compatible with autoregressive generation. Based on this insight, we propose \textbf{LatentOmni}, a cross-modal reasoning framework that interleaves textual reasoning with audio-visual latent states. LatentOmni introduces feature-level supervision to align latent reasoning states with task-relevant sensory features and uses Omni-Sync Position Embedding (OSPE) to maintain temporal consistency between latent audio and visual states. We further construct \textbf{LatentOmni-Instruct-35K}, a dataset of audio-visual interleaved reasoning trajectories for supervising latent-space reasoning. Comprehensive evaluation across multiple audio-visual reasoning benchmarks demonstrates that LatentOmni achieves the best performance among the evaluated open-source models and consistently outperforms the Explicit Text CoT baseline, supporting latent-space joint reasoning as a promising path toward stronger omnimodal understanding.
\end{abstract}

\section{Introduction}
\label{section1_intro}

Information in the real world is inherently multimodal~\cite{hong2025worldsense,zhou2025daily}, and artificial agents must jointly interpret what they see and hear to understand events, causality, and context~\cite{zhu2021deep,baltruvsaitis2018multimodal,zhang2025debiasing,yin2024survey}. Recent multimodal large language models (MLLMs) have made notable progress on audio-visual perception tasks such as captioning and grounding~\cite{chen2025avocado,mm-rlhf,chen2026diadem,shi2025mme,ding2026omnisift,yan2025crosslmm}, yet they remain constrained on reasoning problems that require integrating fine-grained evidence across modalities~\cite{li2025omnivideobench,xing2025echoink}. This gap matters because audio-visual understanding depends not only on recognizing individual signals, but also on reasoning over their temporal and semantic interactions.

\begin{figure}[!t]
  \centering
    \includegraphics[width=\textwidth]{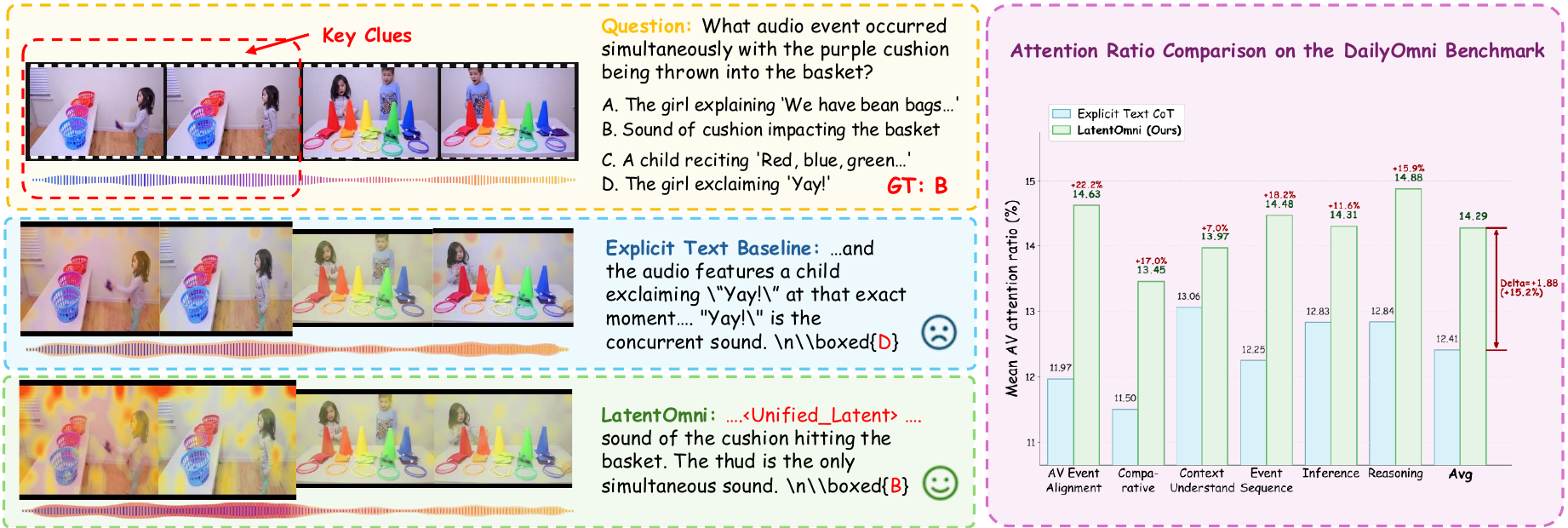}
  \caption{Comparison between LatentOmni and the Explicit Text CoT baseline (detailed in~\ref{4_1_experiment_set}). (Left) Qualitatively, unlike the baseline, LatentOmni accurately anchors on key audio-visual (AV) clues (indicated by heatmaps) to answer correctly. (Right) Quantitatively, it maintains a significantly higher AV token attention ratio across tasks on the Daily-Omni benchmark, ensuring robust grounding of original modalities.}
  \label{fig:intro}
\end{figure}

We identify a key bottleneck in how current MLLMs perform reasoning. Most existing approaches rely on explicit or structured text-based chain-of-thought (CoT)~\cite{wei2022chain,wang2025multimodal,shao2024visual,zhang2023multimodal}, which maps high-dimensional audio-visual evidence into discrete text tokens. This textual bottleneck compresses away temporally aligned details and encourages the model to lean on language priors rather than native sensory evidence during reasoning. As illustrated in Figure~\ref{fig:intro}, pure explicit text CoT therefore tends to under-attend to the original audio-visual inputs, limiting the model's ability to exploit fine-grained cross-modal cues such as temporal synchronization.

We argue that this bottleneck can be mitigated by preserving part of the reasoning process in continuous latent space, where fine-grained audio-visual features are more directly retained than in discretized textual explanations. Motivated by this perspective, we propose \textbf{LatentOmni}, a post-training framework that interleaves textual reasoning with audio-visual latent states in a unified latent space. To keep reasoning grounded in the original modalities, LatentOmni introduces feature-level supervision that aligns latent reasoning states with task-relevant audio-visual segments, encouraging the model to retain and attend to native sensory evidence throughout the reasoning process. To preserve temporal consistency across modalities, we further introduce Omni-Sync Position Embedding (OSPE), which extends the time-aligned multimodal RoPE~\cite{xu2025qwen2} to synchronized latent audio and visual features. Together, these designs enable latent states to serve as a dense bridge between audio, vision, and text while retaining the structural benefits of textual reasoning.

Implementing feature-level supervision within the latent space requires CoT data with pre-annotated, reasoning-relevant audio-visual segments, a form of supervision largely missing from current audio-visual instruction datasets. These datasets typically provide coarse question-answer pairs or textual rationales, without localizing the visual frames and audio intervals that support each reasoning step. To fill this gap, we develop a scalable data curation pipeline featuring audio-video interleaved reasoning trajectory and construct \textbf{LatentOmni-Instruct-35K}, a high-quality dataset specifically tailored for cross-modal reasoning tasks.

As illustrated in Fig.~\ref{fig:intro}, compared to purely explicit CoT reasoning methods, LatentOmni substantially improves attention to the original audio-visual (AV) modalities, particularly on AV alignment tasks. Furthermore, extensive experiments demonstrate that LatentOmni achieves the best results among the evaluated open-source models on all four benchmarks, outperforming both the base model and the explicit text CoT baseline by a clear margin. In brief, our contributions are summarized as follows:
\begin{itemize}
    \item We propose \textbf{LatentOmni}, a novel audio-visual reasoning framework that equips MLLMs with a tailored post-training pipeline to conduct joint reasoning in a unified latent space.
    \item We introduce explicit feature-level supervision in latent space and Omni-Sync Position Embedding (OSPE) to facilitate cross-modal temporal alignment, which efficiently preserves attention to audio-visual modalities and bridges audio-visual with textual semantics.
    \item We develop a novel audio-visual interleaved CoT data synthesis pipeline, and construct \textbf{LatentOmni-Instruct-35K}, a high-quality dataset filling the gap in tailored training data for complex cross-modal latent reasoning.
    \item Our extensive experiments show that LatentOmni substantially outperforms the Explicit Text CoT baseline and achieves state-of-the-art open-source performance on challenging benchmarks, confirming its substantial promise for robust multimodal understanding.
\end{itemize}

\section{Related work}
\subsection{Multimodal Large Language Models Reasoning}
Multimodal Large Language Models (MLLMs) originally aimed to equip LLMs with diverse perceptual capabilities~\cite{girdhar2023imagebind,li2023blip,shi2025mavors,wang2025scone}; however, to tackle complex real-world tasks, research has progressively shifted toward enhancing their reasoning abilities. A prevailing paradigm to achieve this is leveraging explicit chain techniques~\cite{wang2025multimodal,shao2024visual, xie2025audio,ma2025audio, tong2026cof}. By establishing text as the primary semantic bridge for cross-modal integration, these models can effectively decompose complex tasks via natural language rationales~\cite{dong2025insight}. This text-centric reasoning approach has demonstrated encouraging progress in individual visual and audio domains, and has now naturally extended to drive recent omnimodal frameworks like Gemini~\cite{team2023gemini}, Video-LLaMA series~\cite{zhang2023video}, and the Qwen-Omni series~\cite{xu2025qwen2}.


Despite its widespread adoption, recent research reveals that this discrete reasoning paradigm fundamentally constrains complex cross-modal inference~\cite{pham2025multimodal,zhang2025modalities}. Forcing high-dimensional audio-visual signals through a narrow textual bottleneck inevitably causes information loss. Furthermore, this text-centric abstraction results in insufficient attention to raw audio-visual signals. This imbalance leads to sensory detachment and multimodal hallucinations, where generated rationales decouple from the actual underlying evidence~\cite{qian2026cognitive, fang2026seeing}. Although recent tool-augmented approaches (e.g., think with audio, image and video)~\cite{xiong2025thinking,su2025thinking,zhang2025thinking,yan2026act} attempt to mitigate this, they fail to fundamentally resolve the inherent neglect of cross-modal inputs. Consequently, these limitations severely impede the scalability of explicit CoT reasoning~\cite{kancheti2026chain}.

\subsection{Reasoning in Latent Space}

To mitigate the constraints of discrete token generation, recent studies have explored conducting reasoning directly within continuous latent spaces~\cite{goyal2023think,hao2024training, zelikman2024quiet}. As a pioneering work in this direction, Coconut~\cite{hao2024training} bypasses the autoregressive generation of intermediate textual tokens by executing reasoning steps entirely within the model's hidden states. This continuous reasoning paradigm has subsequently been extended to the multimodal domain to better accommodate continuous real-world sensory signals~\cite{bardes2024revisiting}. In this context, current research generally follows two mainstream methodologies: some works design specific training frameworks~\cite{li2025latent, wang2025monet, liu2025reasoning} to optimize reasoning trajectories within the latent space, while others develop training-free inference mechanisms~\cite{li2025latentImplicit} to elicit latent reasoning capabilities directly from pre-trained representations.

Despite these advances, existing latent reasoning methods predominantly focus on pure text or single-modality extensions, such as visual-textual integration~\cite{wang2025monet, li2025latent, li2025latentImplicit, qin2025chain}. The joint comprehension and reasoning of dynamic Audio-Visual (AV) signals within a unified continuous space remains underexplored. Recognizing this gap, our work introduces LatentOmni to extend continuous latent reasoning to omnimodal scenarios, explicitly addressing the temporal and semantic alignment of cross-modal AV integration.

\section{Method}
\begin{figure}[!t]
  \centering
  \includegraphics[width=\textwidth]{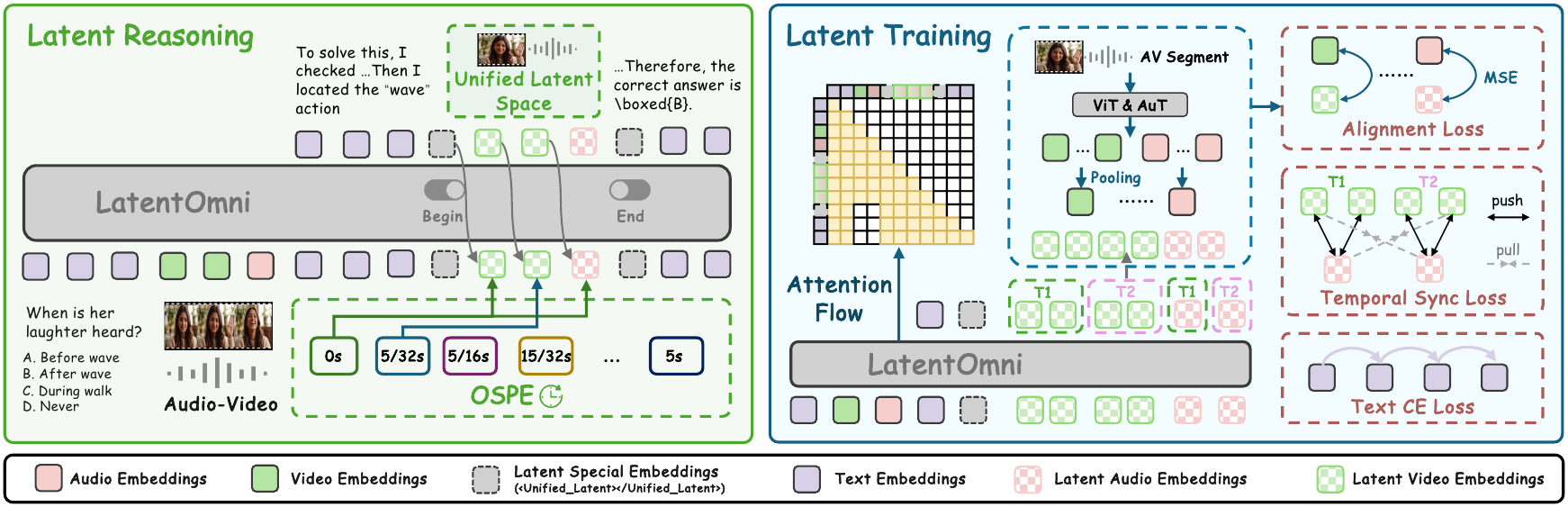}
  \caption{Overview of LatentOmni. Left: the model alternates between textual generation and latent reasoning. Right: training combines text prediction, latent alignment, and temporal synchronization objectives.}
  \label{fig:LatentOmni}
\end{figure}

We present LatentOmni, a post-training framework for audio-visual reasoning in a unified latent space. As illustrated in Fig.~\ref{fig:LatentOmni}, the framework combines interleaved text-latent reasoning, synchronized audio-visual latent representations, a dedicated interleaved reasoning dataset, and training objectives that ground latent states in native sensory evidence. We first describe the reasoning process and latent representation design, then present the data synthesis pipeline and the training objectives.

\subsection{Audio-Visual Latent Reasoning}
Text-only CoT provides useful logical structure, but it is inefficient for revisiting dense audio-visual evidence. LatentOmni therefore alternates between explicit textual deduction and latent reasoning phases that operate directly on continuous audio-visual states. Given encoded visual features $H^v$, audio features $H^a$, and a textual query $H^q$, the model autoregressively generates a hybrid sequence of text tokens and latent states. When it needs to revisit audio-visual evidence, it emits a special token $\mathtt{<Unified\_Latent>}$, which switches decoding from the discrete vocabulary space $\mathcal{V}$ to a continuous latent space $\mathbb{R}^d$. After generating $K$ latent embeddings, we explicitly insert a stop token $\mathtt{</Unified\_Latent>}$ to terminate the continuous reasoning phase and revert to explicit textual generation. The resulting reasoning trajectory is
\begin{equation}
\label{eq:interleaved_sequence}
S = \left[ w_{1:i}, u, z_{1:K}, u', w_{i+1:j}, u, z_{K+1:2K}, u', \dots, a \right],
\end{equation}
where $w$ denotes text tokens, $u$ is the $\mathtt{<Unified\_Latent>}$ trigger, $u'$ is the inserted $\mathtt{</Unified\_Latent>}$ stop token, $z$ denotes continuous latent reasoning states, and $a$ is the final answer. This design keeps text as the scaffold for high-level logic while reserving latent states for evidence-intensive cross-modal reasoning. We analyze the effect of the latent length $K$ in Section~\ref{4_3_ablation_study}.

\subsection{Unified Latent Representation and Temporal Alignment}

A remaining design question is how to represent latent reasoning states while preserving temporal correspondence across modalities. During each latent reasoning phase triggered by $u$, the model generates a sequence of continuous states auto-regressively. At the $k$-th latent step, the latent representation $z_k \in \mathbb{R}^d$ is instantiated as the last-layer hidden state of the transformer backbone prior to the language modeling head (Fig.~\ref{fig:LatentOmni}, left):
\begin{equation}
z_k = \operatorname{LM}_{\theta}^{(L)} \left( H^v, H^a, H^q, S_{<k} \right),  
\end{equation}
where $L$ denotes the number of transformer layers and $S_{<k}$ is the preceding mixed context of text tokens and latent states. Each generated $z_k$ is then fed back as the input embedding for the next latent step, forming a continuous reasoning trajectory of length $K$. We allocate the first $K_v$ positions to visual latents and the remaining $K_a$ positions to audio latents, which lets the model control modality-specific capacity while keeping all latent states in the same continuous space $\mathbb{R}^d$.

Sequential generation, however, creates a mismatch risk: audio and visual latents that refer to the same moment may drift apart positionally. To prevent this, we introduce Omni-Sync Position Embedding (OSPE), which extends the time-aligned multimodal RoPE from Qwen2.5-Omni~\cite{xu2025qwen2} to the unified latent space. OSPE assigns a shared physical timestamp $t$ to temporally corresponding visual frames and audio segments. For a latent feature $h \in \{h^v, h^a\}$ at timestamp $t$, OSPE applies
\begin{equation}
\operatorname{OSPE}(h, t) = h \odot \cos(t \Theta) + \mathcal{R}(h) \odot \sin(t \Theta),
\end{equation}
where $h^v$ and $h^a$ denote latent visual and audio features, $\Theta = \{\theta_i\}_{i=1}^{d/2}$ is the base frequency vector, $\odot$ denotes the Hadamard product, and $\mathcal{R}(\cdot)$ is the block-diagonal rotation matrix over adjacent feature dimensions. By injecting a synchronized positional prior, OSPE aligns sequentially generated latent features that correspond to the same time window, allowing later reasoning steps to attend to temporally consistent cross-modal evidence.

\begin{figure}[!t]
  \centering
  \includegraphics[width=\textwidth]{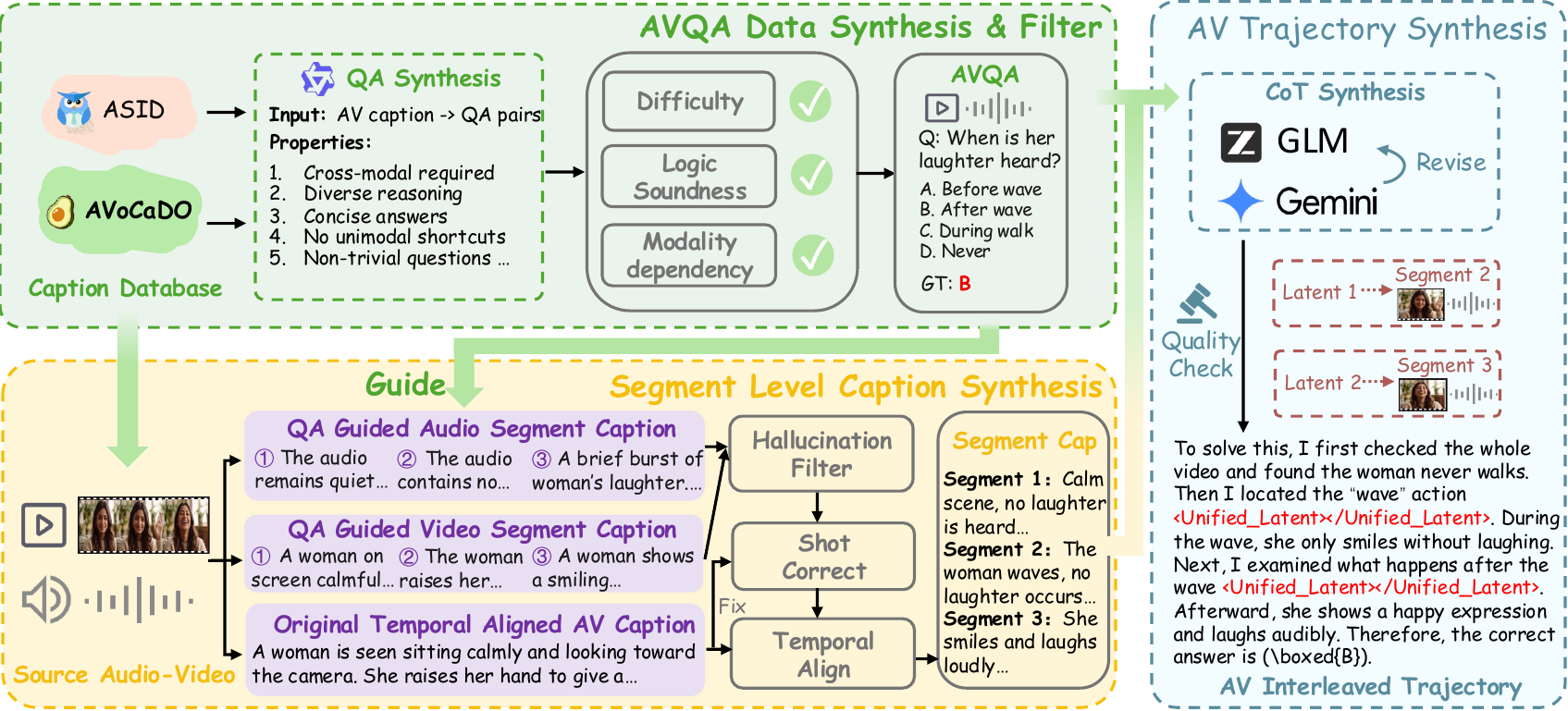}
  \caption{Construction pipeline of LatentOmni-Instruct-35K.}
  \label{fig:data_pipeline}
\end{figure}

\subsection{LatentOmni-Instruct-35K Dataset Construction}
\label{3_3_data_pipeline}
Latent-space reasoning requires supervision beyond standard question-answer pairs: the model must know which local audio-visual evidence should be revisited at each step. Existing datasets rarely provide such segment-grounded interleaved trajectories. We therefore build \textbf{LatentOmni-Instruct-35K} through a three-stage pipeline, shown in Fig.~\ref{fig:data_pipeline}, consisting of AVQA synthesis and filtering, segment-level caption synthesis, and audio-visual interleaved reasoning trajectory synthesis.

\textbf{AVQA Data Synthesis \& Filtering.} We first collect raw samples from two temporally aligned audio-visual caption datasets, ASID~\cite{li2026towards} and AVoCaDO~\cite{chen2025avocado}, and use Qwen3-235B-A22B~\cite{yang2025qwen3} to transform cross-modal captions into preliminary question-answer pairs. During generation, the model is instructed to produce questions that require cross-modal dependency, cover diverse reasoning types, and preserve answer correctness. We then use GLM-4.7~\cite{zeng2025glm} to assign each pair a category and three quality scores: difficulty, logical soundness, and modality dependency. Samples with a total score below 13 are discarded, and the ratio between any two adjacent categories is constrained to be within $3\times$ to avoid severe imbalance. This stage yields a higher-quality AVQA pool with stronger logical rigor and modality coupling. Prompts are provided in Appendices~\ref{app_avqa_sys} and~\ref{app_avqa_classify_filter}.

\textbf{Segment-Level Caption Synthesis.} Each retained sample also needs localized audio and visual evidence. We therefore segment the raw streams by timestamp and generate segment-level descriptions. Because joint audio-visual captions often omit one modality~\cite{chen2025avocado}, we use Qwen3-30B-A3B-Captioner~\cite{yang2025qwen3} to produce separate audio and video captions for each segment. Using the original aligned source captions as references, GLM-4.7 then filters hallucinated descriptions, repairs shot fragmentation, and realigns the audio and video captions in time. The result is a set of segment-level captions that are both locally grounded and cross-modally aligned. Prompts are provided in Appendices~\ref{app_segment_caption_synthesis} and~\ref{app_segment_caption_fusion}.

\textbf{Audio-Visual Interleaved Reasoning Trajectory Synthesis.} Finally, we synthesize full reasoning trajectories from the filtered AVQA pairs and segment-level captions. GLM-4.7 generates reasoning chains that insert explicit markers whenever a step requires a specific audio-visual segment. Gemini-2.5-Flash then audits these trajectories by correcting citation errors and removing redundant or inconsistent branches. After discarding trajectories with major hallucinations or contradictions, we replace the markers with their corresponding audio-visual segments to obtain the final 35K-sample dataset.

\subsection{LatentOmni Training}
\label{3_4_training}

Our training objective must satisfy three requirements simultaneously: preserve temporal correspondence between audio and vision, ground latent states in native sensory evidence, and retain the model's language-generation ability. We therefore perform supervised fine-tuning on LatentOmni using the audio-visual interleaved CoT dataset from Sec.~\ref{3_3_data_pipeline} and optimize three complementary objectives over the hybrid reasoning trajectory.

Before asking the model to reason over joint latent states, we first align synchronized audio and visual evidence in the shared space through a \textbf{temporal synchronization objective ($\mathcal{L}_{\text{sync}}$)}. Given latent visual features $h_t^v$ and audio features $h_t^a$ at matching timestamps $t \in \mathcal{T}$, we optimize a symmetric InfoNCE contrastive loss:
\begin{equation}
\mathcal{L}_{\text{sync}} = - \frac{1}{2 |\mathcal{T}|} \sum_{t \in \mathcal{T}} \left( \log \frac{\exp \left( \operatorname{sim}(h_t^v, h_t^a) / \tau \right)}{\sum_{t'} \exp \left( \operatorname{sim}(h_t^v, h_{t'}^a) / \tau \right)} + \log \frac{\exp \left( \operatorname{sim}(h_t^a, h_t^v) / \tau \right)}{\sum_{t'} \exp \left( \operatorname{sim}(h_t^a, h_{t'}^v) / \tau \right)} \right),
\end{equation}
where $\operatorname{sim}(\cdot, \cdot)$ denotes cosine similarity and $\tau$ is a learnable temperature. This loss pulls together temporally co-occurring audio-visual features while pushing apart asynchronous pairs, thereby establishing a temporally coherent latent space before deeper reasoning takes place.

Temporal alignment alone, however, does not guarantee that latent reasoning remains attached to the source evidence. To counter the language-bound tendency identified in Sec.~\ref{section1_intro}, we additionally ground each auto-regressively generated latent embedding $z_k$ in raw sensory features. For each annotated audio-visual segment, we extract features using the model's visual and audio encoders and compress them into a dense anchor sequence $A = [a_1, \dots, a_K]$, consisting of $K_v$ visual and $K_a$ audio anchors ($K = K_v + K_a$). We use parameter-free L2-norm-weighted pooling for this compression so that salient transient actions and acoustic events are preserved. As reasoning unfolds auto-regressively, each generated state $z_k$ is aligned with its corresponding anchor $a_k$ using a latent alignment loss:
\begin{equation}
\mathcal{L}_{\text{latent}} = \frac{1}{K} \sum_{k=1}^{K} \left\| z_k - a_k \right\|^2_2.   
\end{equation}

Latent supervision should not come at the expense of the model's linguistic priors. We therefore apply a standard \textbf{next-token prediction loss ($\mathcal{L}_{\text{text}}$)} over all discrete tokens in the hybrid sequence. Given a reasoning sequence $S = \{s_1, s_2, \dots, s_L\}$ containing both text tokens and continuous latent states, we compute the auto-regressive cross-entropy loss only on the elements that belong to the vocabulary $\mathcal{V}$:
\begin{equation}
\mathcal{L}_{\text{text}} = - \frac{1}{N_{\text{text}}} \sum_{t=1}^{L} \mathbb{I}(s_t \in \mathcal{V}) \log p(s_t \mid S_{<t}, H^v, H^a, H^q),
\end{equation}
where $\mathbb{I}(\cdot)$ is the indicator function, $N_{\text{text}}$ is the number of discrete tokens (including text reasoning tokens $w$, the trigger token $u$, and the final answer $a$), and $S_{<t}$ denotes the preceding hybrid context. This preserves the model's ability to perform explicit textual deduction while conditioning each token on the interleaved history of text and latent evidence.

The model is optimized end-to-end with the combined objective function:
\begin{equation}
\label{3_4_equation}
\mathcal{L}_{\text{total}} = \mathcal{L}_{\text{text}} + \lambda_1 \mathcal{L}_{\text{latent}} + \lambda_2 \mathcal{L}_{\text{sync}},
\end{equation}
where $\lambda_1$ and $\lambda_2$ are balancing hyperparameters. The final objective jointly balances textual fluency, modality grounding, and temporal alignment, enabling LatentOmni to reason with continuous audio-visual evidence without abandoning the structural benefits of language.

\section{Experiments}
\subsection{Experimental Setup}
\label{4_1_experiment_set}
\textbf{Training.} 
Following the pipeline in Section~\ref{3_4_training}, we train LatentOmni from Qwen2.5-Omni-7B using LatentOmni-Instruct-35K (Section~\ref{3_3_data_pipeline}). We fine-tune the model for 750 steps (2 epochs), so the comparison mainly reflects the effect of the proposed post-training objective rather than a change in backbone scale. Unless otherwise stated, both training and evaluation use a fixed budget of 40 latent tokens, selected by ablating the total token count and the audio-visual allocation ratio. This fixed setting keeps the inference interface identical across examples and avoids per-sample tuning of the latent length. It is also consistent with prior observations that fixed latent budgets are more stable than dynamic schedules in practical reasoning settings~\cite{li2025latent}.

\textbf{Benchmarks.} We evaluate audio-visual joint reasoning on four omnimodal benchmarks that stress complementary capabilities: everyday scenario reasoning (Daily-Omni~\cite{zhou2025daily}), physical and spatial-temporal commonsense (WorldSense~\cite{hong2025worldsense}), cross-modal alignment and question answering (OmniVideoBench~\cite{li2025omnivideobench}), and long-form multi-sensory understanding (LVOmniBench~\cite{tao2026lvomnibench}). This benchmark suite is intended to test whether latent reasoning helps beyond a single data regime: Daily-Omni emphasizes common event understanding, WorldSense tests structured commonsense over time and space, OmniVideoBench contains fine-grained audio-type and video-duration splits, and LVOmniBench stresses sustained reasoning over longer inputs.

\textbf{Baselines.} We organize baselines to match the analysis order in Section~\ref{4_2_main_results}. First, we compare with representative open-source audio-visual MLLMs, including VideoLLaMA2-7B~\cite{cheng2024videollama}, MiniCPM-o-7B~\cite{yao2024minicpm}, VITA-1.5-7B~\cite{fu2025vita}, HumanOmniV2-7B~\cite{yang2025humanomniv2}, Baichuan-Omni-1.5, OmniVinci, and the Qwen2.5-Omni-7B base model~\cite{xu2025qwen2}. Second, we isolate the effect of latent reasoning from text-only reasoning and ordinary fine-tuning under the same backbone. \textbf{Explicit Text CoT} removes all interleaved audio-video segments from LatentOmni-Instruct-35K and fine-tunes Qwen2.5-Omni-7B on strictly textual reasoning trajectories, while \textbf{Vanilla SFT} directly fine-tunes Qwen2.5-Omni-7B on LatentOmni-Instruct-35K without latent-space reasoning. This pair of controls separates three factors that are otherwise easy to conflate: additional instruction data, explicit textual rationales, and continuous audio-visual latent states. Third, we compare with recent visual latent reasoning methods, Monet~\cite{wang2025monet} and LVR~\cite{li2025latent}, under their vision-only setting. We also report proprietary systems, including GPT-4o~\cite{hurst2024gpt}, Gemini-2.0-Flash, Gemini-2.5-Pro~\cite{comanici2025gemini}, and Gemini-3-Pro~\cite{pichai2025new}, as reference points rather than directly controlled baselines.

\begin{table}[!t]
\caption{Performance on four omnimodal benchmarks. Proprietary systems are included as reference points; the best result among open-source models and Qwen2.5-Omni variants is highlighted.}
\label{tab:performance-comparison}
\centering
\begin{tabularx}{\textwidth}{l *{4}{>{\centering\arraybackslash}X}}
\toprule
Method & Daily-Omni & WorldSense & OmniVideoBench & LVOmniBench \\
\midrule
\rowcolor{yellow!10}
\multicolumn{5}{c}{\textbf{\textit{Open-source Models}}} \\
VideoLLaMA2-7B & 35.2 & 25.4 & 29.2 & 27.0 \\
MiniCPM-o-7B & 53.1 & 29.7 & 29.7 & 34.8 \\
VITA-1.5-7B & 44.7 & 36.9 & 30.5 & - \\
HumanOmniV2-7B & 58.5 & 47.1 & 30.5 & 32.3 \\
Baichuan-Omni-1.5 & 50.0 & 43.3 & 30.7 & 32.8 \\
OmniVinci & 66.5 & 48.2 & 32.1 & - \\
Qwen2.5-Omni-7B & 62.9 & 45.4 & 29.3 & 32.0 \\
\quad + Explicit Text CoT & 65.6 & 46.6 & 33.2 & 32.1 \\
\quad + Vanilla SFT & 62.0 & 47.5 & 30.5 & 33.2 \\
\rowcolor{blue!5}
\textbf{LatentOmni} & \textbf{67.4} & \textbf{48.9} & \textbf{35.4} & \textbf{35.1} \\
\midrule
\rowcolor{green!5}
\multicolumn{5}{c}{\textbf{\textit{Proprietary Models (Reference)}}} \\
GPT-4o & 56.5 & 42.6 & - & - \\
Gemini-2.0-Flash & 67.8 & 56.2 & 41.5 & 42.9 \\
Gemini-2.5-Pro & 81.4 & 64.6 & 58.9 & - \\
\bottomrule
\end{tabularx}
\end{table}

\begin{table}[!t]
  \caption{Accuracy comparison on OmniVideoBench. Closed-source systems are reported as reference points; within open-source rows, the \textbf{best} result is highlighted and the second-best is \underline{underlined}. The gain of LatentOmni over the base model is shown in red parentheses.}
  \label{tab:omnivideo_r1_omnivideobench}
  \centering
  \resizebox{\textwidth}{!}{
  \begin{tabular}{lcccccccc}
    \toprule
    \textbf{Method} 
    & \multicolumn{3}{c}{\textbf{Audio Type}} 
    & \multicolumn{4}{c}{\textbf{Video Duration}} 
    & \textbf{Avg.} \\
    \cmidrule(lr){2-4} \cmidrule(lr){5-8}
    & \textbf{Music} & \textbf{Sound} & \textbf{Speech}
    & \textbf{(0,1] min} & \textbf{(1,5] min} & \textbf{(5,10] min} & \textbf{(10,30] min}
    & \\
    \midrule
    \multicolumn{9}{c}{\textit{Closed-source Models}} \\
    Gemini-2.0-Flash & 29.7 & 40.3 & 43.2 & 49.4 & 43.2 & 41.1 & 34.9 & 41.5 \\
    Gemini-2.5-Pro & 38.5 & 57.7 & 61.7 & 57.8 & 64.4 & 55.0 & 55.9 & 58.9 \\
    Gemini-3-Pro & 56.2 & 54.1 & 55.7 & 61.0 & 56.4 & 52.9 & 52.5 & 55.5 \\
    \midrule
    \multicolumn{9}{c}{\textit{Open-source Models}} \\
    VideoLLaMA2-7B & 26.4 & 30.7 & 29.3 & 32.0 & 28.2 & 29.6 & 28.3 & 29.2 \\
    VITA-1.5-7B & 25.3 & 28.6 & 31.5 & 31.3 & 27.4 & 30.6 & \textbf{34.0} & 30.5 \\
    HumanOmniV2-7B & 20.9 & 31.1 & 31.6 & 36.6 & 29.4 & 29.6 & 29.3 & 30.5 \\
    Baichuan-Omni-1.5-7B & 24.2 & \underline{31.3} & 31.4 & 28.9 & 31.8 & 28.4 & \underline{32.4} & 30.7 \\
    Qwen2.5-Omni-7B & 23.1 & 25.3 & 30.7 & \underline{41.6} & 27.4 & 25.3 & 26.7 & 29.3 \\
    \quad + Explicit Text CoT & \underline{30.0} & \textbf{32.0} & \underline{33.9} & 39.4 & \underline{32.7} & \underline{31.0} & 30.7 & \underline{33.2} \\ 
    \rowcolor{gray!20}
    \textbf{LatentOmni} & \textbf{33.3} & 30.2 & \textbf{36.7} & \textbf{45.2} & \textbf{33.2} & \textbf{33.3} & \textbf{34.0} & \textbf{35.4} \textcolor{red}{(+6.1pp)} \\
    \bottomrule
  \end{tabular}
  }
\end{table}


\subsection{Main Results} 
\label{4_2_main_results}
Table~\ref{tab:performance-comparison} summarizes the main results on four omnimodal benchmarks. We report proprietary systems for context, but focus the controlled comparison on open-source models, text-only reasoning variants, and latent reasoning baselines. Overall, LatentOmni achieves the best performance among the evaluated open-source methods on all four benchmarks, supporting the effectiveness of unified latent-space reasoning for audio-visual tasks.

\begin{table}[!t]
\caption{Comparison with recent visual latent reasoning methods on VideoMME under the vision-only protocol used by prior work.}
\label{tab:performance-latent}
\centering
\begin{tabularx}{\textwidth}{l *{4}{>{\centering\arraybackslash}X}}
\toprule
Method & Overall & Short Video & Medium Video & Long Video \\
\midrule
LVR & 36.7 & 39.2 & 36.6 & 34.3 \\
Monet & 51.6 & 52.9 & 56.0 & 46.0 \\
\rowcolor{blue!5}
\textbf{LatentOmni} & \textbf{60.8} & \textbf{70.8} & \textbf{60.5} & \textbf{50.4} \\
\bottomrule
\end{tabularx}
\end{table}

\begin{table}[!t]
  \caption{Ablation of the components of LatentOmni. The \textbf{best} is highlighted.}
  \label{tab:ablation_model}
  \centering
  \resizebox{\textwidth}{!}{
  \begin{tabular}{lcccc}
    \toprule
    Method & Daily-Omni & WorldSense & OmniVideoBench & LVOmniBench \\
    \midrule
    w/o Audio in Latent Space & 65.9 & 47.8 & 33.6 & 31.6 \\
    w/o Visual in Latent Space & 63.5 & 47.2 & 33.5 & 32.1\\
    w/o OSPE & 66.0 & 47.8 & 34.9 & 33.1\\
    w/o $\mathcal{L}_{\text{latent}}$ & 61.0 & 45.2 & 31.8 & 30.2 \\
    w/o $\mathcal{L}_{\text{sync}}$ & 65.9 & 47.1 & 34.0 & 33.1 \\
    Qwen2.5-Omni-7B & 62.9 & 45.4 & 29.3 & 32.0 \\
    \quad + Explicit Text CoT & 65.6 & 46.6 & 33.2 & 32.1 \\
    \quad + Vanilla SFT & 62.0 & 47.5 & 30.5 & 33.2 \\
    \rowcolor{blue!5}
    \textbf{LatentOmni (full)} & \textbf{67.4} & \textbf{48.9} & \textbf{35.4} & \textbf{35.1} \\
    \bottomrule
  \end{tabular}
  }
\end{table}

\textbf{Comparison with Open-Source Models.} LatentOmni consistently improves over existing open-source audio-visual models. Compared with its base model, Qwen2.5-Omni-7B, LatentOmni obtains absolute gains of 4.5\% on Daily-Omni, 3.5\% on WorldSense, 6.1\% on OmniVideoBench, and 3.1\% on LVOmniBench. It also outperforms strong open-source competitors such as OmniVinci and HumanOmniV2-7B on the benchmarks where they report results. The improvement is especially clear on OmniVideoBench, where LatentOmni reaches 35.4\% and surpasses all evaluated 7B open-source models, indicating stronger cross-modal alignment and reasoning.

\textbf{Comparison with Text CoT.} We next compare LatentOmni with text-only and standard fine-tuning variants built on the same base model. Although Explicit Text CoT improves Qwen2.5-Omni-7B, LatentOmni further raises accuracy by 1.8\% on Daily-Omni, 2.3\% on WorldSense, 2.2\% on OmniVideoBench, and 3.0\% on LVOmniBench. Relative to Vanilla SFT, LatentOmni also yields gains on all datasets, with the largest improvements on Daily-Omni (+5.4\%) and OmniVideoBench (+4.9\%). These controlled comparisons suggest that the gain does not come merely from additional instruction tuning or textual rationales, but from preserving reasoning-relevant audio-visual evidence in latent states.

\textbf{Comparison with Latent Reasoning Methods.} We further compare with recent visual latent reasoning methods, LVR and Monet, on VideoMME. Because these methods are vision-centric, we follow a vision-only protocol without audio inputs. As shown in Table~\ref{tab:performance-latent}, LatentOmni achieves the highest overall score (60.8) and leads across short, medium, and long videos. This result suggests that the proposed latent reasoning design remains effective even when evaluated outside the full audio-visual setting.

\textbf{Fine-Grained OmniVideoBench Analysis.} Table~\ref{tab:omnivideo_r1_omnivideobench} provides a more detailed view of cross-modal reasoning behavior. Among open-source methods, LatentOmni achieves the highest average accuracy (35.4\%), improving over the base model by 6.1pp. It leads on music and speech questions, all short-to-medium duration buckets, and ties for the best score on the longest videos ((10,30] min). Compared with Explicit Text CoT, LatentOmni improves the average accuracy by 2.2pp and shows a clear advantage on long-form video reasoning (34.0\% vs. 30.7\% on the longest subset), supporting the benefit of synchronized continuous latent states for sustained audio-visual understanding.

\subsection{Ablation Study}
\label{4_3_ablation_study}
We ablate the main design choices of LatentOmni to identify where the gains come from. Specifically, we examine the modality composition of the unified latent space, the role of OSPE, the latent sequence configuration, and the individual contributions of $\mathcal{L}_{\text{latent}}$ and $\mathcal{L}_{\text{sync}}$. Unless otherwise noted, ablations follow the same evaluation protocol as the main experiments.

\textbf{Component Analysis.} Table~\ref{tab:ablation_model} shows that removing either audio or visual features from the latent space consistently degrades performance, confirming that both modalities contribute to the final reasoning trajectory. Removing OSPE also reduces accuracy on every benchmark (e.g., $67.4 \rightarrow 66.0$ on Daily-Omni and $35.1 \rightarrow 33.1$ on LVOmniBench), supporting the importance of cross-modal temporal alignment. Among the training objectives, $\mathcal{L}_{\text{latent}}$ is the most influential: without it, performance drops sharply to 61.0 on Daily-Omni and 31.8 on OmniVideoBench. Ablating $\mathcal{L}_{\text{sync}}$ yields smaller but consistent losses, indicating that temporal synchronization complements latent grounding rather than replacing it.

\begin{figure}[!t]
  \centering
  \includegraphics[width=\textwidth]{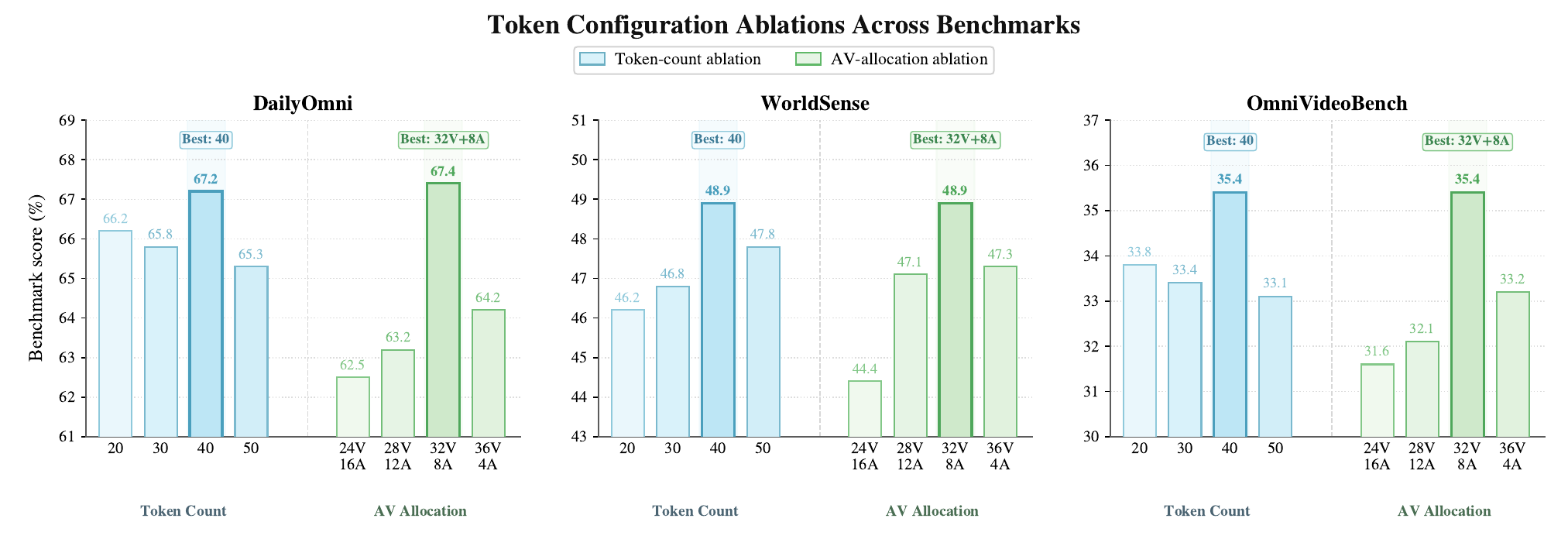}
  \caption{Ablation studies on latent configurations across three benchmarks, specifically evaluating the impact of latent token counts and the allocation ratio of audio and visual latents.}
  \label{fig:Ablation_Hyper}
\end{figure}

\textbf{Impact of Latent Token Configuration.} Figure~\ref{fig:Ablation_Hyper} further studies the length and modality allocation of latent reasoning trajectories. Scaling the total number of latent tokens shows an empirical optimum at 40 tokens: shorter sequences appear to limit representational capacity, while longer sequences add computation without consistent gains. With the length fixed to 40, allocating 32 tokens to visual latents and 8 tokens to audio latents achieves the best overall performance. These results support our default configuration and suggest that audio-visual latent reasoning benefits from a larger visual budget while still requiring a dedicated audio allocation.

\section{Conclusion}
This paper addresses audio-visual reasoning in MLLMs by proposing \textbf{LatentOmni}, a framework that interleaves explicit textual reasoning with synchronized latent audio-visual states. The key idea is to keep intermediate reasoning grounded in native sensory evidence rather than forcing every step through a text-only bottleneck. To this end, we introduce feature-level latent supervision, Omni-Sync Position Embedding (OSPE) for cross-modal temporal alignment, and \textbf{LatentOmni-Instruct-35K} for supervising audio-visual interleaved reasoning trajectories. Across four omnimodal benchmarks, LatentOmni consistently improves over both Qwen2.5-Omni-7B and an Explicit Text CoT baseline, and achieves the best performance among the evaluated open-source models. These results demonstrate the promise of latent-space joint reasoning as a practical and effective path toward more faithful omnimodal understanding.

\medskip

\small
\bibliography{ref}

\newpage

\appendix

\section*{Appendices}
\definecolor{PromptPurple}{RGB}{128,0,128}
\definecolor{PromptPurpleDark}{RGB}{102,0,102}
\newtcolorbox{promptstylebox}[1]{
  enhanced,
  colback=white,
  colframe=PromptPurple,
  boxrule=1.0pt,
  arc=2.5mm,
  left=4mm,
  right=4mm,
  top=4mm,
  bottom=3mm,
  fontupper=\small\itshape,
  title=#1,
  colbacktitle=PromptPurpleDark,
  coltitle=white,
  fonttitle=\bfseries,
  boxed title style={
    arc=1.5mm,
    boxrule=0pt,
    left=4mm,
    right=4mm,
    top=1.2mm,
    bottom=1.2mm
  },
  attach boxed title to top left={xshift=9mm,yshift*=-\tcboxedtitleheight/2},
}

\section{Datasets Details}
\subsection{Caption Database}
\textbf{AvoCaDO} is a newly curated dataset consisting of 107K high-quality, temporally-aligned audiovisual video captions. It emphasizes the temporal orchestration between visual and auditory events, offering semantically rich descriptions paired with precise temporal synchronization. This dataset is specifically designed to enhance temporal coherence, dialogue accuracy, and comprehensive multimodal alignment in audiovisual video captioning tasks.

\textbf{ASID} features a large-scale collection of one million structured, fine-grained audiovisual instruction annotations (ASID-1M). It provides single- and multi-attribute supervision—covering scenes, objects, actions, speech, camera movements, and narrative elements. Curated through an automated verification and refinement pipeline, this dataset is designed to mitigate hallucinations and facilitate highly controllable, reliable, and fine-grained video understanding.

\subsection{AVQA Synthesis Instruction}
\label{app_avqa_sys}
To synthesize high-quality AVQA data in both open-ended and multiple-choice question (MCQ) formats, we design specific instructions. These prompts direct the model to generate question-answer pairs that satisfy strict cross-modal dependencies and adequately reflect the diversity of the source captions. The complete prompts for open-ended QA and MCQ synthesis are illustrated in Fig.~\ref{fig:prompt_qa} and Fig.~\ref{fig:prompt_mcq}, respectively.

\subsection{AVQA Classification and Filtering Instruction}
\label{app_avqa_classify_filter}
To implement the category annotation and quality control described in Section~\ref{3_3_data_pipeline}, we design a joint instruction for GLM-4.7. This prompt directs the model to classify the reasoning type of each preliminary AVQA pair and assess its overall quality, facilitating our threshold-based filtering process. The detailed prompt for classification and evaluation is illustrated in Fig.~\ref{fig:prompt_evaluation}.

\subsection{Segment Level Caption Synthesis Instruction}
\label{app_segment_caption_synthesis}
To synthesize high-quality audio-visual segment-level captions, we design two distinct prompts tailored for the visual and audio modalities. These instructions direct the model to produce separate, concise, and objective descriptions that are strongly correlated with the QA pairs, thereby preventing information omission. The specific prompts for synthesizing observable visual elements and identifiable audio elements are illustrated in Fig.~\ref{fig:prompt_video_caption} and Fig.~\ref{fig:prompt_audio_caption}, respectively.

\subsection{Segment Level Caption Fusion and Refinement Instruction}
\label{app_segment_caption_fusion}
To synthesize comprehensive and cohesive audio-visual captions while resolving narrative fragmentation caused by shot transitions, we design two sequential instructions for caption fusion and refinement. The detailed prompts for the fusion and refinement processes are illustrated in Fig.~\ref{fig:prompt_fuse_caption} and Fig.~\ref{fig:prompt_segment_refinement}, respectively.

\subsection{AV Interleaved Reasoning Trajectory Synthesis Instruction}
\label{app_av_traj_synthesis}
To construct the audio-visual interleaved reasoning trajectories, we design an instruction for GLM-4.7 utilizing the generated AVQA pairs and aligned segment-level captions. This prompt directs the model to logically integrate these multimodal elements into a cohesive, step-by-step reasoning process. The detailed prompt for this trajectory synthesis is illustrated in Fig.~\ref{fig:prompt_traj}.

\section{Implementation Details}
\label{Impletation}
We set the key hyperparameters for our training process as follows: the maximum number of frames per sample (FPS\_MAX\_FRAMES) is capped at 256. For optimization, the learning rate is set to $10^{-5}$, with a warmup fraction of 0.05 to gradually ramp it up at the start of training. The weighting coefficients for the loss terms are configured as $\lambda_1 = 0.005$ and $\lambda_2 = 1.0$. Furthermore, due to limited computational resources, we restrict the batch size to 1, paired with 12 gradient accumulation steps to maintain an adequate effective batch size for stable optimization.

\begin{figure*}[!h]
\centering
\begin{minipage}{\textwidth}
\begin{promptstylebox}{Prompt 1: AV Open-Ended Question-Answer Synthesis}
\textbf{Role.} You are an expert multimodal dataset designer specializing in Audio-Visual Question Answering (AVQA).

\textbf{Input.} Temporally aligned audio and video captions describing synchronized events.

\textbf{Task Description.}
\begin{enumerate}
    \item \textbf{Context Comprehension:} Thoroughly analyze the provided audio and video captions to understand the synchronized multimodal events and their temporal correlations.
    \item \textbf{Open-Ended Generation:} Synthesize exactly \textbf{one} high-quality, open-ended question-answer pair that demands complex, multi-step cross-modal reasoning.
    \item \textbf{Structured Output:} Format the final result strictly as a JSON object containing the question, concise answer, and the specific reasoning type employed.
\end{enumerate}

\textbf{Hard Constraints.}
\begin{itemize}
    \item \textbf{Cross-Modal Information Dependency:} The question must strictly rely on the synthesis of both visual and audio information. It must be logically impossible to deduce the answer using only a single modality.
    \item \textbf{Reasoning Typology:} The generated question must explicitly target a distinct complex reasoning category (e.g., causal reasoning, spatial relations, temporal sequencing, sound-action attribution, or object interactions).
    \item \textbf{Answer Accuracy:} The question and its concise answer (maximum 10 words) must be factually accurate, concrete, and grounded strictly within the provided caption content, with zero external hallucination.
    \item \textbf{Format and Style:} Avoid using object IDs, bounding box labels, timestamps, or raw XML tags in the generated question and answer.
    \item \textbf{Commonsense Integration:} Encourage structural world knowledge when appropriate, such as logically linking visible physical actions with expected environmental acoustics.
\end{itemize}

\textbf{Output Format.} Output raw JSON only. Do not wrap the output in Markdown blocks.

\textbf{Reference JSON Schema.}
{\ttfamily\footnotesize
\begin{verbatim}
{
  "id": "OpenQA_01",
  "modality": "AV",
  "question": "...",
  "answer": "..."
}
\end{verbatim}
}
\end{promptstylebox}
\end{minipage}
\caption{Prompt used to synthesize a single complex, open-ended AVQA pair from temporally aligned audio-visual captions.}
\label{fig:prompt_qa}
\end{figure*}

\begin{figure*}[!ht]
\centering
\begin{minipage}{\textwidth}
\begin{promptstylebox}{Prompt 2: AV Multiple-Choice QA Synthesis}
\textbf{Role.} You are an expert multimodal dataset designer specializing in Audio-Visual Question Answering (AVQA).

\textbf{Input.} Temporally aligned audio and video captions describing synchronized events.

\textbf{Task Description.}
\begin{enumerate}
    \item \textbf{Context Comprehension:} Thoroughly analyze the provided audio and video captions to understand the synchronized multimodal events and their temporal correlations.
    \item \textbf{Multiple-Choice Generation:} Synthesize exactly \textbf{one} high-quality multiple-choice question (MCQ) that demands complex, multi-step cross-modal reasoning.
    \item \textbf{Structured Output:} Format the final result strictly as a JSON object containing the question, four distinct options, the correct answer, and the specific reasoning type employed.
\end{enumerate}

\textbf{Hard Constraints.}
\begin{itemize}
    \item \textbf{Cross-Modal Information Dependency:} The question must strictly rely on the synthesis of both visual and audio information. It must be logically impossible to deduce the answer using only a single modality.
    \item \textbf{Reasoning Typology:} The generated question must explicitly target a distinct complex reasoning category (e.g., causal reasoning, spatial relations, temporal sequencing, sound-action attribution, or object interactions).
    \item \textbf{Distractor Quality:} The MCQ must contain exactly four options. The three incorrect options (distractors) must be plausible, misleading, and reflect reasonable alternative interpretations, but must be definitively incorrect without any ambiguity or subjectivity.
    \item \textbf{Answer Accuracy:} The strictly correct answer must be factually accurate, concrete, and grounded entirely within the provided caption content, with zero external hallucination.
    \item \textbf{Format and Style:} Avoid using object IDs, bounding box labels, timestamps, or raw XML tags in the generated question and options.
\end{itemize}

\textbf{Output Format.} Output raw JSON only. Do not wrap the output in Markdown blocks.

\textbf{Reference JSON Schema.}
{\ttfamily\footnotesize
\begin{verbatim}
{
  "id": "MCQ_01",
  "modality": "AV",
  "question": "...",
  "options": [
    "A. ...",
    "B. ...",
    "C. ...",
    "D. ..."
  ],
  "answer": "B",
  "answer_text": "B. ..."
}
\end{verbatim}
}
\end{promptstylebox}
\end{minipage}
\caption{Prompt used to synthesize a single complex, multiple-choice AVQA pair with grounded distractors and a unique correct answer from temporally aligned captions.}
\label{fig:prompt_mcq}
\end{figure*}

\begin{figure*}[!h]
\centering
\begin{minipage}{\textwidth}
\begin{promptstylebox}{Prompt 3: AVQA Dataset Quality Evaluation and Classification}
\textbf{Role.} You are an expert evaluator and classifier specializing in Multimodal Large Language Models (MLLMs) and Audio-Visual Question Answering (AVQA) datasets.

\textbf{Objective.} Perform TWO tasks based on the provided inputs:
\begin{enumerate}
    \item Objectively evaluate the quality, rigor, and grounding of the provided Question-Answer pair.
    \item Classify the user's question into one specific AVQA category AND determine its primary modality dependency.
\end{enumerate}

\textbf{Input Data.}
\begin{itemize}
    \item {[Standard AV Caption]}: \{AV\_caption\}
    \item {[Question]}: \{question\}
    \item {[Ground Truth Answer]}: \{answer\}
\end{itemize}

\textbf{Task 1: QA Quality Evaluation (1-5 Scale).}
Evaluate the provided QA pair across the following 6 dimensions:
\begin{enumerate}
    \item \textit{Context Utilization \& Relevance (1-5):} Does the question effectively target the provided modality context? (5 = strictly relies on necessary AV information; 1 = ignores context or relies on external general knowledge).
    \item \textit{Question Difficulty (1-5):} How inherently difficult is the question? (5 = highly complex, multi-step reasoning or nuanced integration; 1 = simple, shallow factual lookup).
    \item \textit{Deductive Requirement (1-5):} Does answering the question require genuine logical deduction from the observations? (5 = requires deep step-by-step inference; 1 = pure parroting or trivial text matching).
\end{enumerate}

\textbf{Task 2: Question Classification \& Modality.}
First, classify the question into EXACTLY ONE of the following 10 categories:
(1) Audio-Visual Joint Perception, (3) Action \& Behavior Recognition, (3) Spatial Layout Understanding, (4) Temporal Sequence Understanding, (5) Attribute Comparison \& Change, (6) Counting \& Quantification, (7) Emotion \& Atmosphere Perception, (8) Semantic Content Summarization, (9) Logical Relation Reasoning, (10) Intention \& Outcome Prediction.

Second, determine the primary modality dependency. Choose EXACTLY ONE:
\begin{itemize}
    \item AV-Strong: Requires logically combining visual and auditory cues.
    \item Video-Strong: Relies primarily on visual information.
    \item Audio-Strong: Relies primarily on auditory information.
\end{itemize}

\textbf{Output Format.} Output ONLY a valid JSON object. Do not include markdown code blocks, conversational text, or explanations outside the JSON structure.

\textbf{Reference JSON Schema.}
{\ttfamily\footnotesize
\begin{verbatim}
{
  "evaluation": {
    "context_utilization": <int, 1-5>,
    "question_difficulty": <int, 1-5>,
    "deductive_requirement": <int, 1-5>
  },
  "classification": {
    "category_id": <int, 1-10>,
    "category_name": "<string>",
    "modality_dependency": "<AV-Strong|Video-Strong|Audio-Strong>",
    "confidence": <float, 0.0-1.0>,
    "reasoning": "<string, 1-2 sentences>"
  }
}
\end{verbatim}
}
\end{promptstylebox}
\end{minipage}
\caption{Prompt used to evaluate the intrinsic quality and modality dependency of synthesized AVQA pairs against standard captions.}
\label{fig:prompt_evaluation}
\end{figure*}

\begin{figure*}[!h]
\centering
\begin{minipage}{\textwidth}
\begin{promptstylebox}{Prompt 4: Segment Level Caption Synthesis (Video)}
Given the provided QA pair, provide a CONCISE yet complete visual-only description of the video segment that contains relevant information to answer the question. Limit the description to no more than five sentences and avoid redundant details. Describe only directly observable visual elements: setting, people, actions, objects, and camera movement. Do not infer mood, intent, genre, cultural style, or add interpretation, and strictly avoid speculation and evaluative language.
\end{promptstylebox}
\end{minipage}
\caption{Prompt used to synthesize a concise video caption focusing exclusively on observable visual elements guided by a specific QA pair.}
\label{fig:prompt_video_caption}
\end{figure*}

\begin{figure*}[!h]
\centering
\begin{minipage}{\textwidth}
\begin{promptstylebox}{Prompt 5: Segment Level Caption Synthesis (Audio)}
Given the provided QA pair, provide a CONCISE yet complete audio-only description of the segment that contains relevant sound information to answer the question. Limit the description to no more than five sentences and avoid redundant details. State only clearly identifiable sound sources (e.g., music, instruments, environmental noises). If speech is present, accurately report the speaker and the spoken content. Do not infer mood, intent, genre, or cultural style, and strictly avoid speculation and atmospheric language.
\end{promptstylebox}
\end{minipage}
\caption{Prompt used to synthesize a concise audio caption detailing identifiable sounds and speech relevant to a specific QA pair.}
\label{fig:prompt_audio_caption}
\end{figure*}

\begin{figure*}[!h]
\centering
\begin{minipage}{\textwidth}
\begin{promptstylebox}{Prompt 6: Segment Level Caption Fusion}
You are tasked with fusing the visual caption and audio caption into a single, coherent narrative based on the video content. Follow these strict rules:

1. Preserve every single sentence from both the visual caption and audio caption exactly as they appear.

2. Do NOT omit or delete any sentence in any way.

3. You may reorder the sentences (from both captions) to create a logical and temporally accurate sequence that reflects the video’s events.

4. Ensure the integrated narrative flows naturally in time with the video, aligning visual actions with corresponding sounds or spoken content.

Verify before responding: Did I include every sentence from both captions?
\end{promptstylebox}
\end{minipage}
\caption{Prompt used to fuse the segment-level video caption and audio caption.}
\label{fig:prompt_fuse_caption}
\end{figure*}

\begin{figure*}[!h]
\centering
\begin{minipage}{\textwidth}
\begin{promptstylebox}{Prompt 7: Segment Caption Refinement}
\textbf{Task:} Refine a fragmented video segment caption using the full video caption as a reference.\vspace{0.5em}

\textbf{Inputs:}
\begin{itemize}
    \item \textbf{[Full Caption]:} [full\_AV\_caption]
    \item \textbf{[Segment Caption]:} [segment\_AV\_caption]
\end{itemize}\vspace{0.5em}

\textbf{Instructions:} Compare the [Segment Caption] with the [Full Caption] to fix shot fragmentation. Resolve incomplete actions, abrupt cuts, or missing subjects caused by rigid segmentation by accurately contextualizing actions that span across boundaries. Do not describe events outside the segment's timestamp. Keep the refined caption concise, strictly objective, and output ONLY the refined text without any formatting tags or explanations.
\end{promptstylebox}
\end{minipage}
\caption{Prompt used to refine fragmented segment captions by cross-referencing full captions to address shot fragmentation and maintain temporal constraints.}
\label{fig:prompt_segment_refinement}
\end{figure*}

\begin{figure*}[!h]
\centering
\begin{minipage}{\textwidth}
\begin{promptstylebox}{Prompt 8: AV Interleaved Reasoning Trajectory Synthesis}
\textbf{Role.} You are an intelligent audio-visual analysis expert with advanced
perception capabilities. Your task is to answer the user's question by rigorously
analyzing the audio and video content step by step.

\textbf{Critical Context.} You will be provided with specific segments and detailed descriptive content. \textbf{You must treat this detailed content as your own direct sensory perception (sight and hearing).} You are NOT analyzing a text; you are describing the raw video and audio you are "watching".

\textbf{Strict Prohibitions.}
\begin{itemize}
    \item Never mention words such as ``caption'', ``text'', ``description'',
    ``reference'', ``provided info'', or ``input''.
    \item Never say ``according to the description'' or ``the text says''.
    \item Never discuss discrepancies between visual or audio evidence and text; instead,
    state the perceived event as the ground truth.
\end{itemize}

\textbf{Reasoning Instructions.}
\begin{enumerate}
    \item Construct a step-by-step reasoning chain that actively decides when source
    segments must be revisited.
    \item When a segment is needed, cite it using the format {[Segment n]}.
    \item Cite at least one segment and at most three segments in total, favoring the
    earliest or most decisive evidence.
    \item Continue the reasoning without explicitly referring back to segment numbers in
    the narrative text.
\end{enumerate}

\textbf{Final Answer Requirement.} 

End the response with a concise boxed answer in the
format \textbackslash boxed\{your final answer here\}, with no extra
commentary afterward.

\textbf{Reference Output Pattern.}

Reason -> [Segment n] -> Reason -> ... -> \textbackslash boxed\{final answer\}
\end{promptstylebox}
\end{minipage}
\caption{Prompt used to synthesize interleaved reasoning trajectories with explicit
segment citations and concise grounded observations.}
\label{fig:prompt_traj}
\end{figure*}

\clearpage

\section{Case Study}
\begin{figure}[!h]
  \centering
  \includegraphics[width=\textwidth]{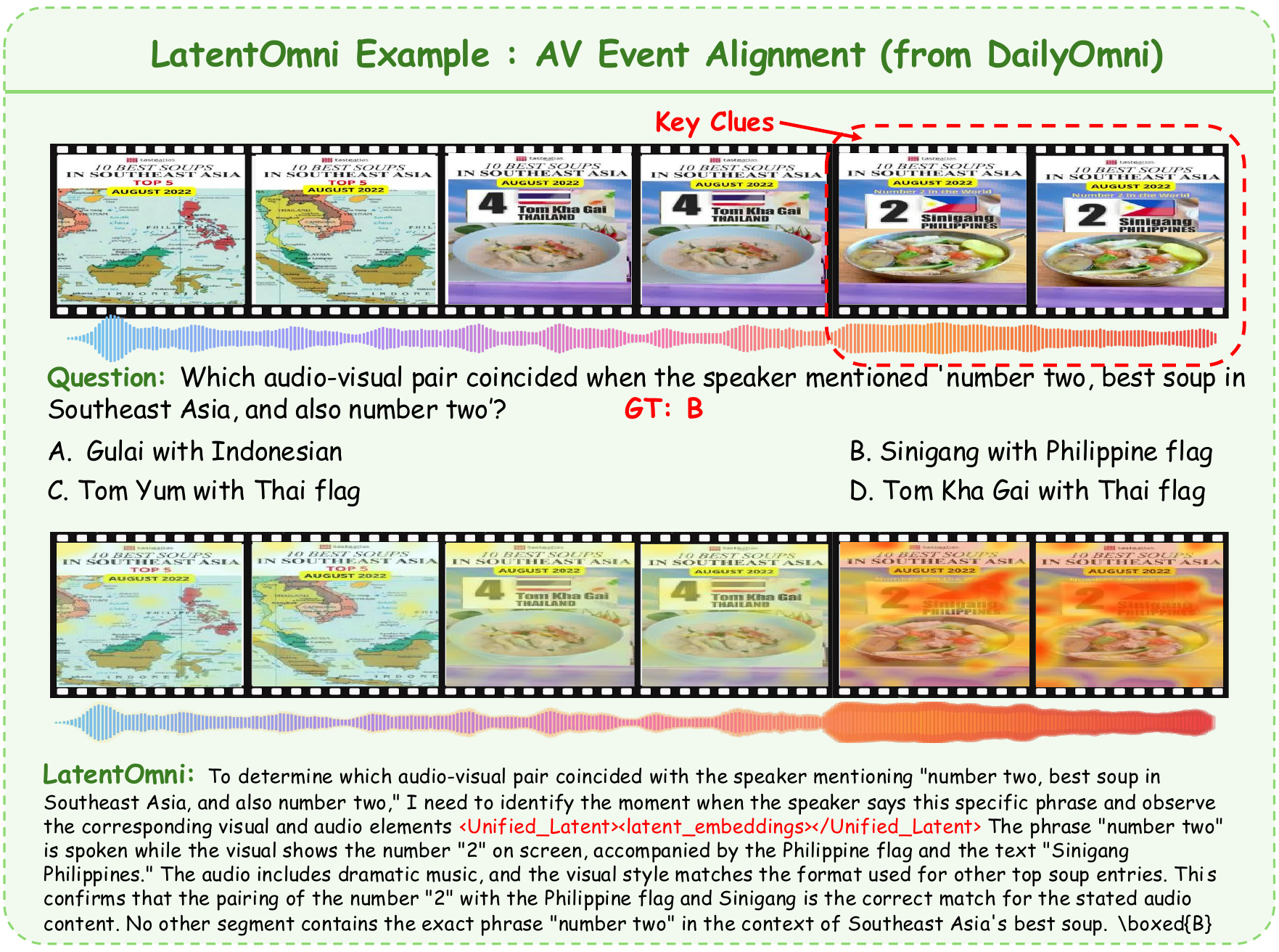}
  \caption{LatentOmni example: AV Event Alignment. LatentOmni accurately anchors task-relevant audio-visual frames within the latent space. As demonstrated by the attention visualization, deeper colors indicate higher attention weights precisely localized on the key multimodal clues.}
  \label{fig:case1}
\end{figure}

\begin{figure}[!h]
  \centering
  \includegraphics[width=\textwidth]{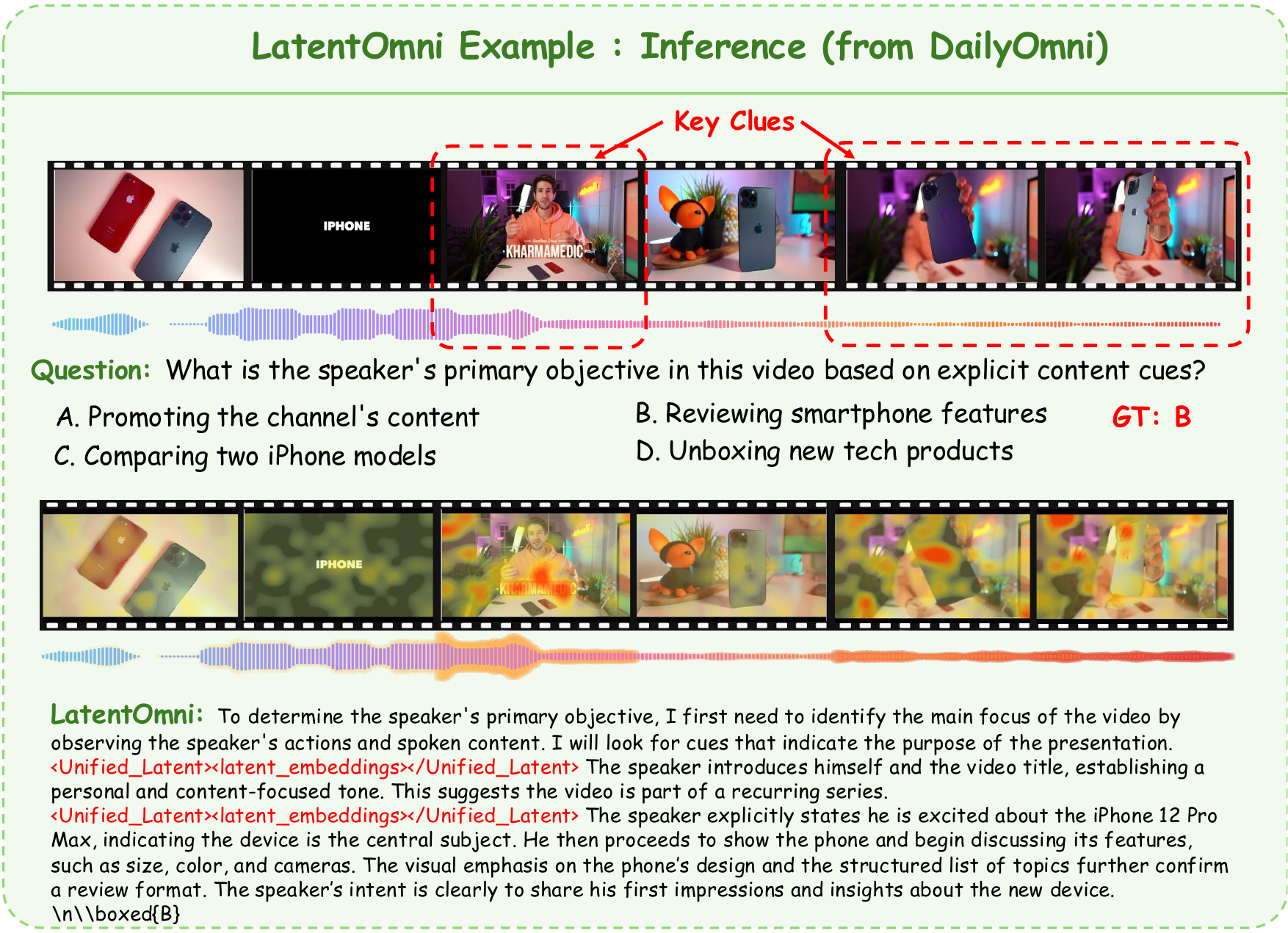}
  \caption{LatentOmni example: Inference. LatentOmni accurately anchors task-relevant audio-visual frames within the latent space. As demonstrated by the attention visualization, deeper colors indicate higher attention weights precisely localized on the key multimodal clues.}
  \label{fig:case2}
\end{figure}

\begin{figure}[!h]
  \centering
  \includegraphics[width=\textwidth]{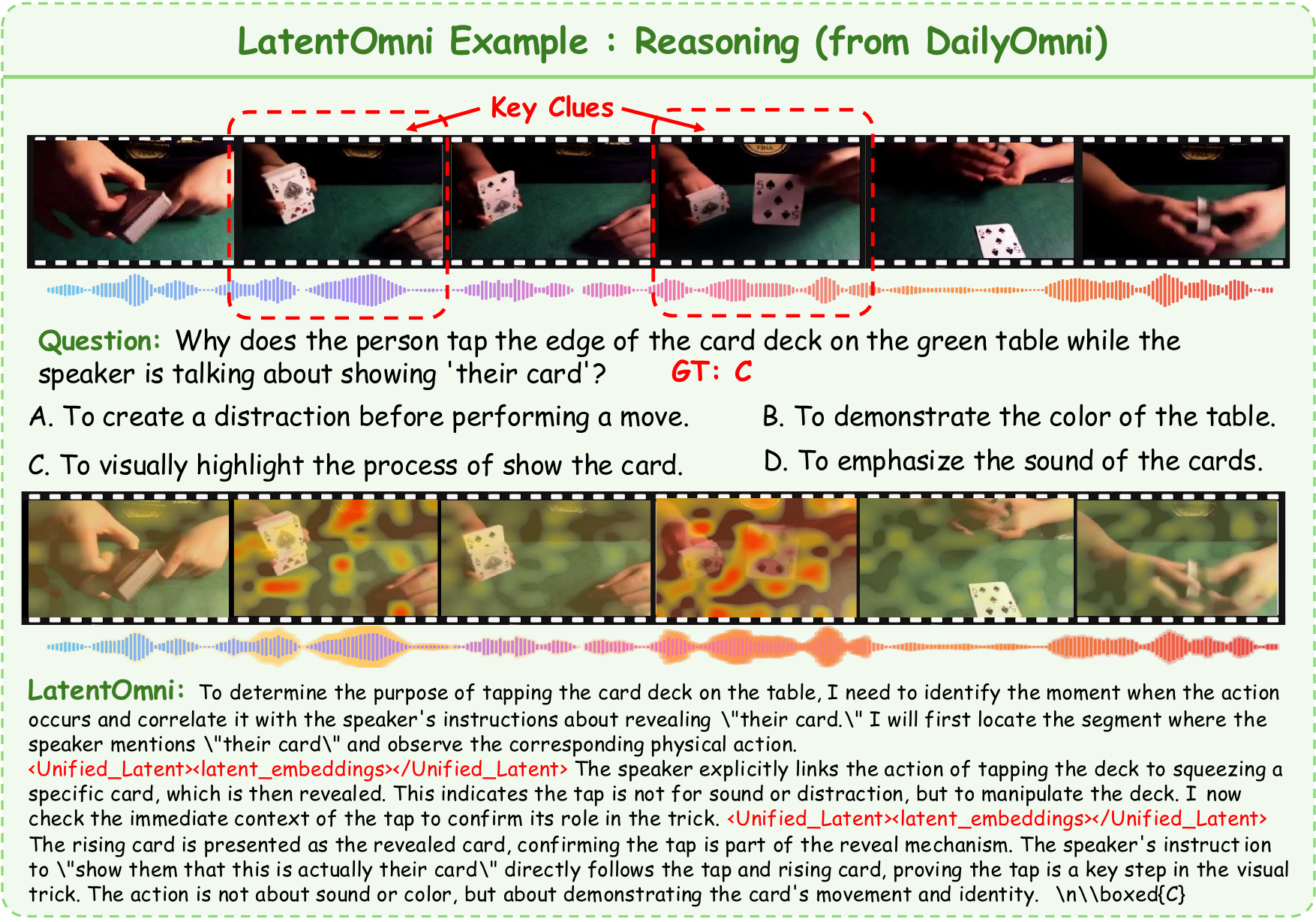}
  \caption{LatentOmni example: Reasoning. LatentOmni accurately anchors task-relevant audio-visual frames within the latent space. As demonstrated by the attention visualization, deeper colors indicate higher attention weights precisely localized on the key multimodal clues.}
  \label{fig:case3}
\end{figure}

In this section, we present examples from DailyOmni benchmark to demonstrate the reasoning capabilities of LatentOmni across diverse audio-visual tasks. For clarity, we abstain from projecting the generated latent embeddings into the discrete language space, as this would yield uninterpretable tokens. Instead, we represent the latent reasoning segments using <Unified\_Latent><latent\_embeddings></Unified\_Latent>. The selected examples encompass three representative scenarios: AV Event Alignment (Figure~\ref{fig:case1}), Inference (Figure~\ref{fig:case2}), and Reasoning (Figure~\ref{fig:case3}).

To further analyze the model's intrinsic reasoning process within the latent space, we concurrently visualize the attention maps between the latent reasoning tokens and the original audio-visual inputs alongside the generated outputs. These visualizations intuitively illustrate how the latent states dynamically track and anchor to fine-grained multimodal evidence during generation. These qualitative analyses highlight the model's capability to preserve and reason over continuous multimodal semantics throughout the inference process.

\section{Limitation}
\label{limitation}
While LatentOmni establishes a robust framework for unified latent reasoning across visual, auditory, and textual modalities, it inherently shares a common boundary with current state-of-the-art multimodal systems regarding modality coverage. Real-world environments are fundamentally more complex, encompassing a broader spectrum of sensory and control signals, such as 3D spatial representations, tactile physics, and motor action commands. Currently, mapping these extended physical and interactive signals into a single unified latent space remains an open challenge for the community. In future work, we aim to explore the expansion of our latent semantic bridge to accommodate a wider array of heterogeneous modalities, ultimately taking a step towards a more comprehensive and embodied omni-modal reasoning system.

\clearpage


\newpage

\end{document}